\documentclass[letterpaper, 10 pt, conference]{ieeeconf}
\IEEEoverridecommandlockouts                              % This command is only
\usepackage{cite}
\usepackage{amsmath,amssymb,amsfonts}
\usepackage{algorithmic}
\usepackage[ruled,vlined,linesnumbered]{algorithm2e}
\usepackage{graphicx}
\usepackage{textcomp}
\usepackage{xcolor}
\usepackage{todonotes}
\usepackage{nicefrac}
\usepackage{comment}
\usepackage{wrapfig}
\usepackage{multirow}
\usepackage{makecell}
\usepackage{subfig}
\usepackage[outercaption]{sidecap}
\usepackage{caption}
\usepackage{bm,times}
\usepackage{marginnote}
\usepackage{siunitx}
\usepackage{url}
\def\BibTeX{{\rm B\kern-.05em{\sc i\kern-.025em b}\kern-.08em
    T\kern-.1667em\lower.7ex\hbox{E}\kern-.125emX}}
\begin{document}

\title{ILoSA: Interactive Learning of Stiffness and Attractors \\
\author{Giovanni Franzese$^{*\dagger}$, Anna M\'esz\'aros$^\dagger$, Luka Peternel, and Jens Kober}
%{\footnotesize \textsuperscript{*}Note: Sub-titles are not captured in Xplore and should not be used}
\thanks{This research is funded by  European Research
Council Starting Grant TERI ``Teaching Robots Interactively'', project reference 804907.}
\thanks{The authors are with Cognitive Robotics, Delft University of Technology, The Netherlands (e-mail: g.franzese, l.peternel, j.kober@tudelft.nl, a.meszaros@student.tudelft.nl).}
\thanks{$^*$Corresponding Author $^\dagger$These authors contributed equally to this work.} 
%\thanks{This work has been submitted to the IEEE for possible publication. Copyright may be transferred without notice, after which this version may no longer be accessible}
% \thanks{1. \textbf{This work has been submitted to the IEEE for possible publication. Copyright may be transferred without notice, after which this version may no longer be accessible.}}
%\thanks{Identify applicable funding agency here. If none, delete this.}
}
\maketitle

\begin{abstract}
Teaching robots how to apply forces according to our preferences is still an open challenge that has to be tackled from multiple engineering perspectives.
This paper studies how to learn variable impedance policies where both the Cartesian stiffness and the attractor can be learned from human demonstrations and corrections with a user-friendly interface. The presented framework, named ILoSA, uses Gaussian Processes for policy learning, identifying regions of uncertainty and allowing interactive corrections, stiffness modulation and active disturbance rejection. 
The experimental evaluation of the framework is carried out on a Franka-Emika Panda in four separate cases with unique force interaction properties: 1) pulling a plug wherein a sudden force discontinuity occurs upon successful removal of the plug, 2) pushing a box where a sustained force is required to keep the robot in motion, 3) wiping a whiteboard in which the force is applied perpendicular to the direction of movement, and 4) inserting a plug to verify the usability for precision-critical tasks in an experimental validation performed with non-expert users. 
\end{abstract}
%\todo[inline]{The abstract also has to be modified}

\section{Introduction}
Robots have long been a tool for efficiently carrying out repetitive or mundane tasks. As of late, more robotic applications are targeted towards interacting with varying and unknown environments in order to aid people in daily tasks. Quite often, the exact behaviour required for interacting with such environments cannot be directly modelled or is simply too complex to do so. However, people already possess intuitions on how to interact with the world around them and can transfer this knowledge. In this direction, learning through demonstration has become increasingly popular for teaching robots familiar yet complex tasks in an intuitive manner \cite{argall2009survey}.

Learning is especially handy for manipulation tasks, which come with the requirement of exerting a certain degree of force. 
%in order to complete the task. 
The goal of a manipulation operation is not only to perform a desired trajectory but to learn the desired force that the robot has to exercise on its environment in order to accomplish the desired goal. Different methods exist for %teaching and 
controlling the robot to perform contact tasks, from the use of force control, hybrid position-force control \cite{raibert1981hybrid}, as well as impedance control methods \cite{hogan1984impedance}. In addition, when robots are coexisting with humans, it is crucial to consider that for safe interaction the robot should limit the force to the minimum required, as well as be compliant when interacting with elements of the environment that are not the target of the manipulation.

Out of these methods, impedance control is best suited for achieving such behaviour, since force control would generate dangerous and unstable accelerations when in free movement \cite{kober2015learning}, while impedance control would only converge to the nearby attractor. Furthermore, when using impedance control, for a safer and user-friendly interaction of the robot with the environment, the trajectory execution has to be performed in a feedback/reactive manner, not in feed-forward. This avoids the accumulation of error between the attractor and end-effector positions with the consequent generation of undesired high interaction forces and/or accelerations. 
\label{sec::lisarm}
\begin{figure}[!t]
    \centering
    \includegraphics[width=0.48\textwidth, trim=0.95cm 0.3cm 1.5cm 0,clip]{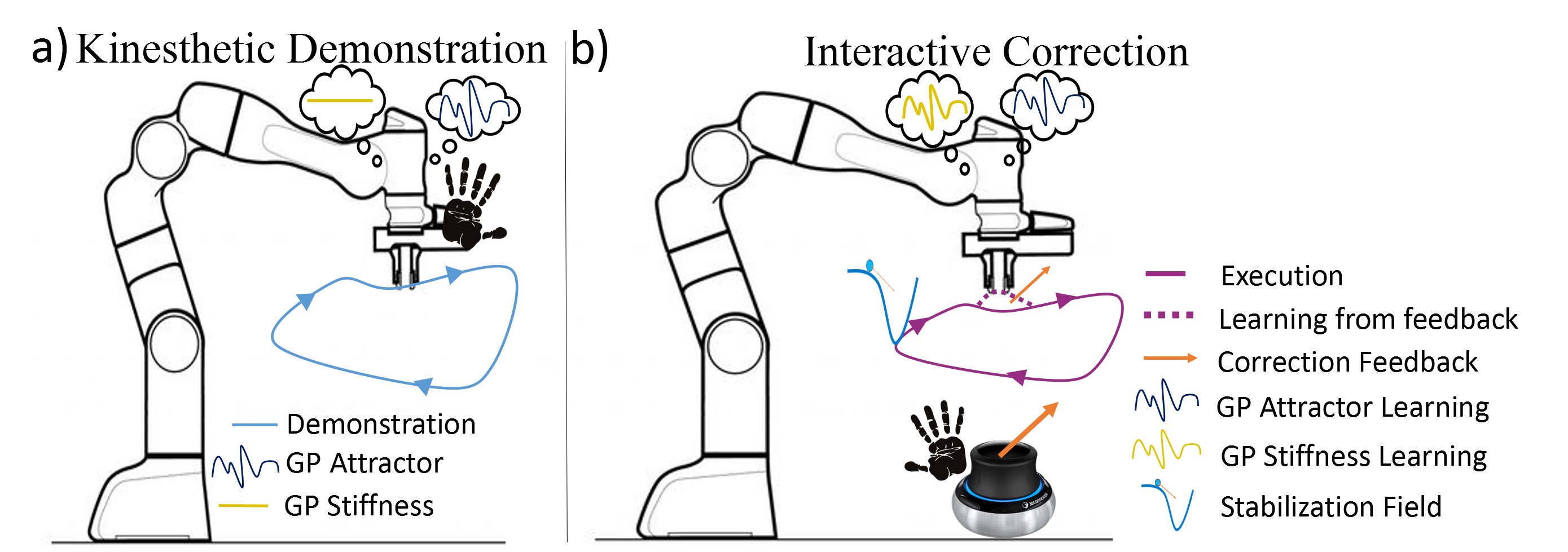}
    \caption{Overview of the ILoSA framework}
    \label{fig:lisarm}
    \vspace{-0.5cm}
\end{figure}

In the presented framework, both the desired attractor position and the desired stiffness are learned as a function of the robot position. This is done by using a nonlinear feedback policy that is learned from kinesthetic demonstrations and teleoperated corrections.
More importantly, because the robot learns from a human demonstration, an estimation of the epistemic uncertainty of the policy is necessary for a safe execution of the trajectory. For this purpose, the Gaussian Process Regression (GPR) provides a parameter-free learning method that enables a good generalization in the neighborhoods of the demonstration while providing information on the confidence level of the corresponding prediction. This in turn can be utilised to make the robot more compliant in states the robot did not visit before, thus avoiding undesired and dangerous behaviours. 
To summarise, the motivation of this paper is to establish whether learning both attractor and stiffness policies in a reactive formulation with a GPR allows the performance of manipulation tasks while ensuring a safe interaction between the robot and its environment by exploiting the information of model confidence. Additionally, we introduce a new update rule for a Gaussian Process (GP) policy in order to allow data and time-efficient learning from teleoperation corrections after the initial kinesthetic demonstration, and to automatically allocate the feedback as attractor or stiffness modulation. Also, we investigate the concept of adaptive disturbance rejection with the use of a stabilization prior based on a force field proportional to the gradient of the GP variance manifold. 

\section{Related Work}
\label{releted_works}
%The impedance controller is designed to take the reference position as an input. 
The impedance controller reacts proportionally to the distance from the desired reference position. This enables the robot to follow a trajectory when in free space or to apply a force when in contact with an object. This dual property is utilised within the scope of the developed framework.

In order to automatically learn policies to complete complex tasks, one of the common approaches is to apply reinforcement learning (RL) algorithms such as Policy Improvement with Path Integrals (PI$^2$) \cite{kalakrishnan2011learning,buchli2011learning,hazara2016reinforcement}, Natural Actor-Critic \cite{kim2010nac}, and multi-optima policy search \cite{calinon2013compliant}. RL methods, however, tend to take a long time to achieve a desired performance level. Additionally, in order to train a policy through RL, an adequate cost function is needed for which a good understanding of the task mechanics is required.

A faster alternative is learning fromin order toion \cite{kober2015learning}, which can additionally be augmented with incremental learning \cite{petrivc2018accelerated} in order to improve a demonstrated policy. Previous works have shown that while using an impedance controller it is possible to learn varying stiffness profiles in order to carry out force interaction tasks \cite{abu2018force,rozo2013learning} as well as to allow compliance in areas outside of force interaction \cite{buchli2011learning}. %Variable impedance control has also been used to learn a progressing automation of a task \cite{kastritsi2018progressive}, however, it homogeneously modifies the stiffness during demonstrations as a function of their consistency and not according to the robot's position by learning from user corrections. 

For the purpose of learning an interaction with the environment, feedback from a human operator may be needed in the form of corrections in the action-space \cite{perez2020interactive}. HG-Dagger \cite{kelly2019hg} proposes to allow the user to take control and provide local demonstrations online which are then aggregated onto the existing database. However, the aggregation operation does not employ a data efficient rule for updating the old database according to the current corrections. 

Inspired by the exploitation of model confidence of active learning \cite{silver2012active}, we learned an overall policy that remains compliant in regions of uncertainty for the purpose of safety. Regression based on Gaussian Processes has proven to be a viable approach towards achieving this form of behaviour in \cite{fanger2016gaussian} where the uncertainty information was utilised in order to allocate leadership in the form of compliance to interacting agents in a bi-manual task. 
On the data efficiency side, PILCO \cite{deisenroth2011pilco} successfully employed GPs in RL for learning the system model and using the information about the uncertainty for a faster search for policy optimization. Analogously, Interactive Learning of Stiffness and Attractors (ILoSA) uses the information of the model uncertainty for a data efficient update of the policy with user corrections, for stiffness modulation in uncertain regions and for active rejection of disturbances with a stabilization function.  

In the literature, other probabilistic methods were applied for modeling the stochasticity of movement primitives \cite{paraschos2013probabilistic} 
and inferring the desired trajectory by conditioning the model on the chosen goal.
Furthermore, methods like Gaussian Mixture Models (GMMs) showed a successful application in inferring the desired manipulability ellipsoid from the consistency of multiple demonstrations %\cite{calinon2018robot}
in combination with a geometry-aware controller \cite{jaquier2018geometry}. Similarly, the same method was employed for the fusion of an imitation policy with a stabilization prior \cite{pignat2019bayesian} reducing the problem of the covariate shift. However, these probabilistic approaches do not allow the interactive correction of the policy after the provided demonstration and do not design the stabilization prior as a function of the model confidence, contrary to ILoSA.

Moreover, we propose to teach the desired force through automatic inference on the increase of the attractor distance or stiffness only from user corrections. Combining this with a novel data efficient update rule and the exploitation of the model confidence provided a user-friendly way of teaching force tasks as described in the following sections.

\section{Framework: ILoSA}

\label{method}

ILoSA employs two main teaching modalities: \textbf{kinesthetic demonstration} and \textbf{teleoperation feedback}, see Fig.~\ref{fig:lisarm}. The first is used for initializing the policy for the desired dynamics of the end-effector. This policy is then executed in the second modality, whereby the user can provide online corrections to the policy.

The aim of the learned policy is to affect two particular aspects of the impedance control: the attractor distance $\Delta\bm{x}$, and the stiffness of the end-effector $\bm{K_s}$.  Briefly, in a Cartesian impedance control \cite{hogan1984impedance}, the end-effector dynamics are modelled in the form of a mass-spring-damper system
\begin{equation}
    \bm{\Lambda}(\bm{q})\ddot{\bm{x}} = \bm{K}_s\Delta \bm{x}-\bm{D}\dot{\bm{x}} + \bm{f}_{ext},
\end{equation}
where $\bm{\Lambda} (\bm{q}) $ is the physical system's Cartesian inertia matrix, 
%$\bm{K}_s$ is a diagonal matrix with the desired stiffness in the principal directions, 
$\bm{D}$ is the corresponding critical damping matrix, and 
%$\Delta \bm{x}$ is the attractor distance. 
$\bm{f}_{ext}$ are the external forces.
%The aforementioned policy consists of multiple GP models, trained to take the 3D Cartesian position as input, and the desired attractor distance and stiffness along each of the three principal directions as output.
In the proposed framework, the controlled 3-D vector $\Delta \bm{x}$ and anisotropic diagonal stiffness matrix $\bm{K}_s$ are the mean values of GPRs, conditioned on the current Cartesian position of the robot.

In the initialization of the policy, following the kinesthetic demonstration, the hyper-parameters of the GP models are optimized for maximizing the expectancy of the predicted attractor distance of the provided demonstrations. The same parameters are then used to initialize GP models of the stiffness in the three principal directions, however choosing a non-zero mean of $K_\mathrm{mean}$ in each direction.
In case a force sensor is installed on the end-effector, the stiffness could be initialized proportionally to the recorded external force. 
Our goal is to show that even without a force sensor, the stiffness can be initialized to a base value and the desired deviations can be learned with the interactive human corrections.

ILoSA additionally incorporates two safety features. The first is a stabilization prior which ensures robust control. The second is a modulation function which pulls the stiffness down to zero in regions of high uncertainty. These two aspects will be explained further in the course of this section.

In the following subsections, details are reported on how the GP learns from the demonstration and corrections (Sec. \ref{subsec:GPcorrection}), how the directional feedback is spread between attractor and stiffness (Sec. \ref{subsec:directionalFB}), and how a stabilization prior (Sec. \ref{subsec:stabilisationPrior}) and stiffness modulation (Sec. \ref{subsec:stiffnessModulation}) are obtained as a function of the process variance.

\subsection{Interactive Learning with Gaussian Processes}

\label{subsec:GPcorrection}

\begin{algorithm}[t]
\SetAlgoLined
\DontPrintSemicolon
\textbf{a) Kinesthetic Demonstration(s)} \\
\While{Trajectory Recording}{ 
{Receive($\bm{x}_t \xrightarrow{} \bm{\xi}$)} \\
 $\Delta \bm{x}^d(\bm{x}_{t-1})=\bm{x}_t-\bm{x}_{t-1}$\\

 }
Train(GPs) \\
\textbf{b) Interactive Corrections} \\
\KwData{$\bm{\xi}, \Delta \bm{x}^d, \bm{K}_s^d$} 
\While{Control Active}{
{Receive($\bm{x}$)} \\
$[\Delta \bm{x}, \Sigma$] = GP$_{\Delta \bm{x}}(\bm{x})$ \\
$\bm{K}_s =$ GP$_{\bm{K}_s}(\bm{x})$ \\
\If{Received Human feedback}{
    $[\Delta \bm{x}_\mathrm{inc}, \bm{K}_{s\_\mathrm{inc}}]$=\textbf{Interpret}(\emph{feedback},~$\Delta \bm{x}, \bm{K}_s$)\\
    \eIf{$\Sigma > \Sigma_{Threshold}$}{{Append($\bm{x} \xrightarrow{} \bm{\xi}, \Delta \bm{x}+\Delta \bm{x}_\mathrm{inc} \xrightarrow{} \Delta \bm{x}^d, \bm{K}_s+\bm{K}_{s\_\mathrm{inc}} \xrightarrow{} \bm{K}_s^d$)}}{{Correct}($\Delta \bm{x}_\mathrm{inc} \xrightarrow{} \Delta \bm{x}^d, \bm{K}_{s\_\mathrm{inc}} \xrightarrow{} \bm{K}_s^d$)}
   {Fit(GPs)}
    }

    $\Delta \bm{x} =$GP$_{\Delta \bm{x}}(\bm{x})$\\
    $\bm{K}_s = $GP$_{\bm{K}_s}(\bm{x})$\\
    % $K_s =$ Modulation(GP$_{K_s}(\bm{x})$, $\Sigma$) \\
    $\bm{f}_\mathrm{stable}=-\alpha\nabla \Sigma$ \\
    $[\Delta \bm{x}, \bm{K}_s]$=\textbf{Modulation}($\Delta \bm{x}, \bm{K}_s, \bm{f}_\mathrm{stable}, \Sigma$) 

{Send}($\Delta \bm{x}, \bm{K}_s)$\\}

 \caption{ILoSA}
 \label{algo::lisarm}
\end{algorithm}
A GP is a non-parametric regression method that provides the means for inferring prediction and epistemic uncertainty with a clear mathematical expression. 
The two equations that govern the mean and the variance of the process are
\begin{equation}
\label{eq::GP}
    \mu (\bm{x}) = \bm{k_*}(\bm{\xi},\bm{x})^\top(\bm{K}(\bm{\xi},\bm{\xi})+\sigma^2_n\bm{I})^{-1} \bm{y}= \bm{A}(\bm{\xi}, \bm{x}) \bm{y},
\end{equation}
\begin{equation}
    \Sigma = k(\bm{x},\bm{x})-\bm{k}_{*}(\bm{\xi},\bm{x})^\top(\bm{K(\bm{\xi},\bm{\xi})}+\sigma^2_n\bm{I})^{-1}\bm{k}_{*}(\bm{\xi},\bm{x}),
\end{equation}
where $k$ is the variance of the single evaluation point $\bm{x}$, $\bm{k}_{*}$ is the covariance between $\bm{x}$ and the training inputs $\bm{\xi}$, $\bm{K}$ is the covariance matrix of the training inputs, $\sigma^2_n$ is the variance of the Gaussian noise of the training points, and $\bm{y}$ denotes the training outputs \cite{Rasmussen2005}. Both $k$ and $\bm{k}_{*}$  as well as $\bm{K}$ are a function of the kernel and its hyper-parameters used to incorporate prior knowledge in the process. 
After the kinesthetic demonstration(s) (Fig.~\ref{fig:lisarm}-a), 
the hyper-parameters are automatically determined through Expectation Maximisation with the L-BFGS method.
We are going to distinguish between \textit{Training} (l.~6 of Alg.~\ref{algo::lisarm}), when the optimization of the hyperparameters is performed, and \textit{Fitting}, when the hyperparamenters are kept constant (l.~19 of Alg.~\ref{algo::lisarm}).
That is, when the interactive corrections are provided through teleoperation with the human-in-the-loop, the hyper-parameters are kept invariant because the correlation between samples (and validity of the same kernel) can also be considered invariant. 

The interactive correction in ll.~13-20 of Alg.~\ref{algo::lisarm} summarizes how ILoSA exploits the use of the uncertainty measure for understanding if the robot is in a new unvisited region (l.~14). This requires to add the corrective sample to the database. Otherwise, it determines how to spread the correction on all existing samples that are correlated with the current end-effector position without adding additional samples. Thus, the update rule of the database (l.~17) is
\begin{equation}
\label{eq::update}
\mu + \epsilon_{\mu}=\bm{A}(\bm{\xi}, \bm{x})(\bm{y}+\bm{\epsilon_y}) \Rightarrow
\bm{y}_\mathrm{new}=\bm{y} + \bm{A}(\bm{\xi}, \bm{x})^+ \epsilon_\mu,
\end{equation}
where $\bm{A}(\bm{\xi}, \bm{x})^+$ is the pseudoinverse of $\bm{A}$ and $\epsilon_\mu$ is the correction provided at $\bm{x}$. % (either $\Delta x_{inc}$ or $K_{s\_inc}$ depending on the dataset being corrected). 
$\bm{A}^+$ can be seen as a selector that automatically modulates how much the correlated elements in the database should be modified for matching the user's desired corrections. This rule was applied for correcting the attractor distance, stiffness or both according to the interpretation of the human feedback, see Sec. \ref{subsec:directionalFB}.

This way of spreading corrections on the database showed to be more user-friendly, as well as time and data efficient. As also in other research works \cite{bobu2020less}, having contradictory, incremental or multimodal data can generate a bias of the predicted solution towards the more frequent samples. When doing policy correction, this can result in the unobservability of the effect of feedback and in the user's frustration. The proposed update rule resolves this problem and proved to also be effective in rapidly adjusting mistaken corrections provided in the previous policy roll-outs without their accumulation as noise in the database.    
\subsection{Directional Feedback Interpretation}
\label{subsec:directionalFB}
Our aim is to make the teaching as simple as possible for non-skilled users without any knowledge about robot control.
As the name of the framework suggests, the goal is to learn the attractor and stiffness for the robot end-effector. Related works like \cite{ajoudani2012tele} and \cite{peternel2018robotic} already investigated how to teleoperate the robot while also teaching the desired stiffness ellipsoid of the end-effector or recent works like \cite{michel2021bilateral} infers the modulation as a function of external forces. Contrary to this, our goal is to be able to infer the stiffness not by explicitly labelling it but through the same teleoperated feedback intended for correcting the direction of movement without the use of any expensive device and as a function of the robot position. Accordingly, the main idea is to enable the user to incrementally correct the dynamics of the end-effector. 
The attractor distance increment $\Delta x_\mathrm{inc}$ is obtained as the teleoperated input \emph{feedback};
% \begin{equation}
%     \Delta x_{inc}=feedback
% \end{equation}
however if $|\Delta x + \Delta x_\mathrm{inc}| \geq \Delta_\mathrm{lim}$ or $ K_s > K_\mathrm{mean}$ the stiffness change $K_{s\_\mathrm{inc}}$ is obtained by solving the equation 
\begin{equation}
    (K_s+K_{s\_\mathrm{inc}}) \Delta_\mathrm{lim}=K_s |\Delta x + \Delta x_\mathrm{inc} |.
\end{equation}
up to the saturation limit. 
It is important to note that the mentioned operations are carried out axis-wise.

This approach not only simplifies the feedback modality but also facilitates the teaching of force tasks with abrupt discontinuities. For example, in the scenario of cable unplugging, a closer attractor with a higher stiffness helps to prevent the ``recoil" effect upon object separation. Similarly, if we are pushing a heavy box, the limitation of the attractor distance bounds the robot velocity in case the contact point with the box is lost (see Fig.~\ref{fig:box_removed}). This allows a safer coexistence of the robot in anthropocentric environments. % \cite{haddadin2009safeRobots}.

Finally, in the interactive session, for the purpose of explicitly labelling the desired goal point with zero velocity and high stiffness, a further teleoperation input 
was employed. 
\subsection{Stabilizing Attractive Field}
\label{subsec:stabilisationPrior}
\begin{figure*}
    \centering
    \includegraphics[width=0.95\textwidth]{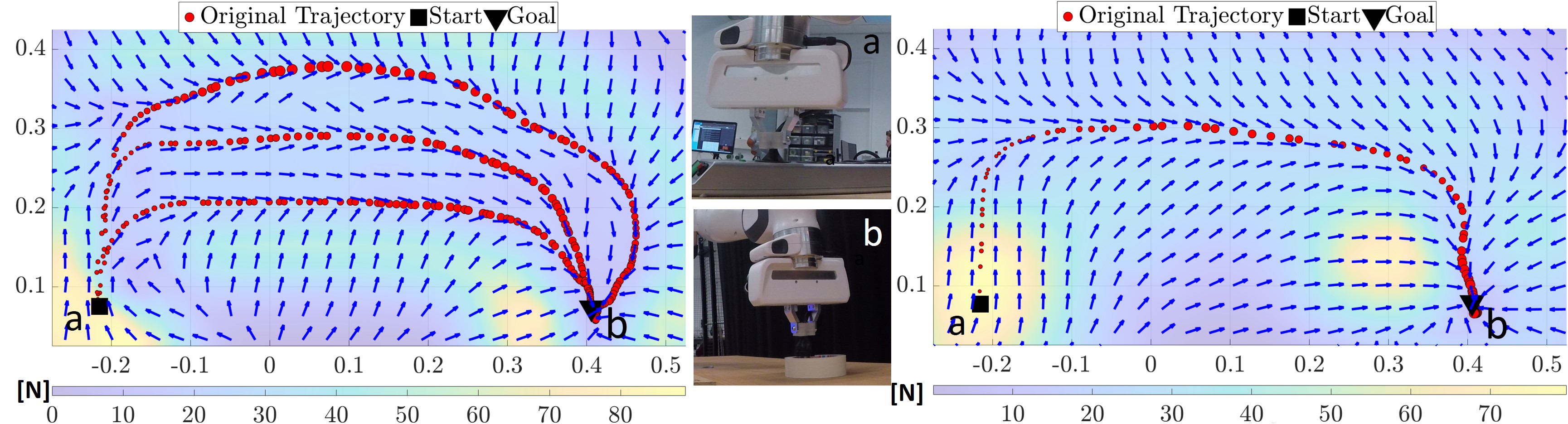}
    \caption{Example of attractor fields for unplugging with multiple demonstrations (left) and a single demonstration (right).}
    \label{fig:plugSingleMultiple}
\end{figure*}

External forces can lead the robot end-effector in previously unvisited regions of the workspace where the extrapolation of the desired $\Delta \bm{x}$ and $\bm{K}_s$ can have high uncertainty and lead to dangerous and undesired dynamics of the robot. This problem, known as covariate shift, is common when applying Behavioural Cloning, 
and some solutions like DART \cite{laskey2017dart} or (HG-)DAgger \cite{kelly2019hg} investigated the injection of noise in the supervised policy execution in order to lead the robot in unvisited regions and collect a database in a larger portion of the environment.
This technique could also be applied in the Interactive Corrections segment (Fig.~\ref{fig:lisarm}-b), however collecting many correction points can be time consuming and highly data inefficient.

As an alternative, we want to exploit the information of the model variance and its continuous differentiability for modelling how to reject external disturbing forces, i.e., not related with the manipulation operation. Intuitively, we can imagine the variance manifold as a hyper-plane with a furrow that is generated in proximity of the labelled regions of the workspace, as we do when we create the circuit for a marble race on the beach. In the absence of external disturbances, the end-effector would lay in the regions of minimum variance and move inside there. However, the robot should reject forces that are leading its motion to a region of uncertainty, proportionally to the rate of change of the same measure. Equivalently, when the external forces are not disturbing the motion anymore, ideally, the robot should go back in regions where the predictions have higher confidence. It is similar to adding a gravitational term in the variance manifold inducing the end-effector to always ``fall" into regions of minimum variance, as a marble would come back on track when disturbed by any collision. The implementation of this stabilization prior is straightforwardly a force field that is proportional to the gradient of the variance manifold according to: 
\begin{equation}
    \bm{f}_\mathrm{stable}(\bm{x})=-\alpha \nabla \Sigma=\alpha \left( 2 \bm{k}_{*}^\top(\bm{K}+\sigma^2_n\bm{I})^{-1}\frac{\partial\bm{k}_{*}}{\partial \bm{x}}\right),
\end{equation}%
where $\bm{x}$ is the evaluation point and $\alpha$ is an automatically modulated constant according to a maximum allowed force, which ensures that $\bm{f}_\mathrm{stable}$ is never higher than the set threshold. Since the kernel is a function of a set of hyperparamenters that are fitted for the different tasks, this turns into a different rejection behaviour according to the fitted model. For example, if the model gets uncertain faster, it will act with a stronger force because it does not want to go in regions where it would not know how to act.  Additionally, the use of this prior also results in another interesting behaviour when multiple demonstrations are provided. In the regions of lower overlapping of the demonstrations we can imagine a broader furrow in the variance manifold compared to regions of highly overlapped demonstrations. In the first case, there is a larger track where the ``marble" can move before reaching the borders and getting pulled down; on the contrary, in the second case the narrow furrow forces the robot to stay closer to the overlapping demonstrations. This different behaviour of reacting to external disturbances can be interpreted as adaptive disturbance compliance of the robot along the lines of 
\cite{jaquier2018geometry}%\cite{calinon2010learning}
, where higher variance in the demonstration results in higher robot affordance and vice-versa. 

\subsection{Stiffness and Attractor Modulation}
\label{subsec:stiffnessModulation}
Finally, before sending the desired attractor and the stiffness to the robot, we want to make sure to spread the effect of $\bm{f}_\mathrm{stable}$ as stiffness and attractor modulation in order to respect the constraint of having a limited attractor and to obtain the desired force with an increase of stiffness. The desired force is $\bm{f}= \bm{K}_s \Delta \bm{x}+ \bm{f}_\mathrm{stable}$, so until the maximum attractor distance is reached $\Delta \bm{x}= \bm{f}/\bm{K}_s$, otherwise the stiffness is also modulated according to $\bm{K}_s= \bm{f} / \Delta_\mathrm{lim}$.

Additionally, when the robot is in a position where the uncertainty approaches the maximum, it is safer to pull the robot stiffness down to zero, rather than the predicted mean value of the GPR, according to 
\begin{equation}
    \bm{K}_s=\bm{K}_s %\odot
    \left(\frac{1-\nicefrac{\Sigma}{\Sigma_\mathrm{max}}}{1-\theta}\right)  \text{      when      } \nicefrac{\Sigma}{\Sigma_\mathrm{max}}> \theta,
\label{eq::modulation}
\end{equation}
where $\Sigma_\mathrm{max}$ is the variance of the unconditioned GP with the defined kernel, and $\theta$ is the relative uncertainty threshold. These two operations are summarized in l.~23 of Alg.~\ref{algo::lisarm}.
 Thanks to the property of distance-based kernels of having the prediction to vanish in high-uncertainty regions, and the modulation of the stiffness, the risk of moving in unknown regions of the workspace with possible undesired behaviours is mitigated. Moreover, this behaviour results in the robot stopping with high compliance and can be seen as a non-verbal request of teaching, or repositioning into regions closer to the demonstration. Finally, this property also circumvents the issue of variable stiffness instability \cite{kronander2016stability} \cite{ferraguti2013tank} making the growing oscillations around the nominal trajectory unobtainable. This would ensure a safe interaction with the user and the environment.

\section{Validation Experiments}

In order to validate our approach we carry out experiments on four different manipulation tasks, each with its own variations, intended to test the different aspects of ILoSA. 
\begin{table*}[ht]
    \parbox{0.7\textwidth}{
    \centering
    \caption{Performance in Unplugging}
    \begin{tabular}{|c||c|c||c|c||c|c||c|c|}
        \hline
         & \multicolumn{2}{c||}{\textbf{Demo Time [s]}} & \multicolumn{2}{c||}{\textbf{Feedback Time [s]}} &  \multicolumn{2}{c||}{\textbf{Data Efficiency [\%]}} & \multicolumn{2}{c|}{\textbf{Goal Error [m]}}\\
         \cline{2-9}
         & Single & Multiple & Single & Multiple & Single & Multiple & Single & Multiple \\
         %\Xhline{4\arrayrulewidth}
         \hline
        \textbf{Max} & 8.00 & 28.27 & 3.47 & 2.13 & 97.06 & 97.06 & 0.016 & 0.030 \\
        \hline
        \textbf{Mean} & 6.73 & 23.84 & 1.96 & 1.61 & 95.36 & 96.10 & 0.009 & 0.014 \\
        \hline
        \textbf{Min} & 5.67 & 21.40 & 1.27 & 1.40 & 92.86 & 95.65 & 0.003 & 0.008 \\
        \hline
    \end{tabular}
    \label{tab:plugSvsM}
    }
    \parbox{0.29\textwidth}{
    \centering
    \caption{Effect of Stabilization}
    \begin{tabular}{|c||c|c|}
        \hline
         & \multicolumn{2}{c|}{\textbf{Goal Error [m]}} \\
         \cline{2-3}
         & Without Prior & With Prior \\
         \hline
        \textbf{Max} & 0.756 & 0.040 \\
        \hline
        \textbf{Mean} & 0.337 & 0.033 \\
        \hline
        \textbf{Min} & 0.073 & 0.019 \\
        \hline
    \end{tabular}
    \label{tab:plugGPvsGrad}
    }
\end{table*}
The first involves removing a plug from its socket and bringing it to a specified goal (Sec.~\ref{sec::unplugging}). 
In this scenario, we test the effect of the number of demonstrations in broadening the variance furrow. Additionally, the effect of the stabilisation prior for rejecting disturbances is analysed by carrying out the control, in one case with the prior active, and in another without it, all while injecting a randomised force disturbance.
The second scenario is pushing a box to a goal (Sec.~\ref{sec::pushingbox}) and observing the handling of contact loss. %While impedance control already ensures that the acceleration does not increase progressively upon contact loss, ILoSA enables the limitation of the observed velocities by limiting the maximum attractor distance. In contrast, if the force were to be achieved by increasing the attractor distance while maintaining a low constant stiffness, these velocities could be noticeably greater. 
 To test this, an ablation study was carried out. 
In a further experiment, a periodic perpetual movement scenario in the form of cleaning a whiteboard is analysed (Sec.~\ref{sec::cleanwhiteboard}). This scenario brings with it the additional challenge that the desired attractor position is located behind the board. Due to the lack of a force sensor, this cannot be inferred from the demonstrations. Instead, user corrections are required for the robot to learn to exert the required force on the board for successfully cleaning it. An additional challenge was addressed to ILoSA in this scenario: validate the flexibility of altering a taught behaviour to new situations. To showcase this, an obstacle was placed along the original trajectory, limiting the possible height of the motion. The aim was then to provide corrections such that the robot would perform the task while also avoiding the obstacle. 
Lastly, we evaluated whether the framework could be utilised for more precision-critical tasks (Sec.~\ref{sec::plugging}). For this, we carried out a validation experiment wherein the robot was taught to insert a plug into a socket by non-expert users.

% All of the experiments excepting the second scenario are carried out by the authors five separate times. For the experiments in the second scenario which involved non-expert participants, the participants performed the task twice. The first was a trial round in order to get familiar with the teleoperation device before the second, official trial round. 

\begin{wrapfigure}{r}{0.12\textwidth}
    \centering
    \includegraphics[width=1\linewidth, trim=0.15cm 0.1cm 0.2cm 0.1cm, clip]{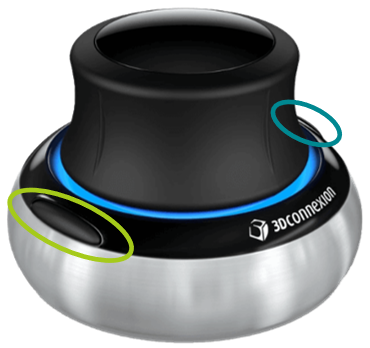}
    \caption{\\SpaceNavigator}
    \label{fig:spacenav}
\end{wrapfigure}
For our experiments we utilise the 7 DoF Franka-Emika Panda with an impedance controller and a ROS communication network for the online update at \SI{100}{\Hz} of the attractor and stiffness using the ILoSA framework.
% Initially, the control frequency was \SI{15}{\Hz} and a majority of the experiments presented in this paper were carried out with this control frequency. The code had since been optimised to enable higher control frequencies, easily achieving a frequency of \SI{100}{\Hz} which was utilised for the experiments of the second scenario.
A 3Dconnexion SpaceNavigator (see Fig.~\ref{fig:spacenav}) was used for providing teleoperation feedback, whereof one of the two buttons (seen circled in Fig.~\ref{fig:spacenav}), was used for explicitly marking the desired goal position as noted at the end of Sec.~\ref{subsec:directionalFB}.
% For all of these experiments excepting the task of inserting a plug, the following parameters were chosen within ILoSA; a mean stiffness of $K_{mean} = \SI{600}{\newton\per\metre}$ with the stiffness limited to $K_{max}=\SI{2000}{\newton\per\metre}$, a maximum attractor distance of \SI{0.04}{\metre} along each principal axis, the epistemic uncertainty threshold for adding new points to the database set to $\Sigma_{Threshold} = 0.2 \Sigma_{max} $, and the threshold for modulating the stiffness $tr = 0.99$. The squared exponential kernel was selected within the GP models. In the case of plugging, the only change to the parameters was the limitation of the stiffness to $K_{max}=\SI{4000}{\newton\per\metre}$, and the maximum attractor distance to \SI{0.02}{\metre}.

For quantifying data efficiency, we compute the ratio between the amount of corrections that result in a modification of the existing database and the total amount of provided feedback inputs. The feedback time was computed as the amount of time explicitly spent providing corrective inputs.

A video of the learning and execution of the tasks can be found attached to this paper \footnote{\url{https://youtu.be/MAG-kFGztws}}. 

\label{experiment}

\subsection{Unplugging}
\label{sec::unplugging}
Two variations of this scenario were performed. In the first, three separate demonstrations were carried out with different heights of the trajectory towards the goal. In the second, a single demonstration was provided. 
%In this scenario, 
The primary focus of the corrections is placed on the successful unplugging as well as reaching the goal within a tolerance of \SI{3}{\cm}. A standard type F plug was used for which specific 3D-printed gripper jaws were designed to ensure a firm grip throughout the interaction. The interaction commences from the point in which the robot is already gripping the plug.  

Fig.~\ref{fig:plugSingleMultiple} visualises examples of the resulting attractor fields generated from $K_s\Delta x$ for both single and multiple demonstrations. As expected, the highest forces are exerted at the beginning, during the unplugging. Instances of moderate forces can be observed leading towards the trajectories. In particular, for the single demonstration, moderate forces are close to the demonstration itself and are, in fact, directed towards the demonstrated trajectory. For the multiple demonstrations these moderate forces are primarily present outside the region of demonstration. This is attributed to the larger variance furrow, which reduces the effect of the stabilisation prior in the demonstrated region, and in turn enables the robot to move more freely within the region when perturbed.

The results regarding the precision in reaching the goal indicate that for both variations, the robot was able to successfully complete its task. Slightly larger errors in the case of multiple demonstrations can be seen. This can, however, be attributed to the variations in the final positions provided during the multiple demonstrations. The time spent giving corrections to complete the task successfully was similar between the two experiments with an average time of \SI{1.96}{\second} for the single demonstration and an average time of \SI{1.61}{\second} for the case with multiple demonstrations. For additional details, refer to Table~\ref{tab:plugSvsM}. For both experiment variations, the majority of feedback inputs did not increase the size of the database, showing in both cases a high data efficiency of more than $95\%$ on average.

In the tests with randomised perturbations, the disturbance was sampled for each of the three axes from a normal distribution $\mathcal{N}(10,5)$\si{\newton} each 0.2 s.
%at 1/3 of ILoSA's update frequency. 
Here, the benefit of the stabilisation prior is clear. When using the stabilisation prior, the error from the goal was on average ten times lower than when the prior was not present.
When using the prior, despite the perturbations, the robot remained close to the goal, with the highest observed error being \SI{4}{\cm}, indicating high robustness. Furthermore, when the prior was not present, the robot had diverged in 3 of the 5 trials and was unable to reach the vicinity of the goal. Additional details are presented in Table~\ref{tab:plugGPvsGrad}.

\subsection{Pushing a Box}
\label{sec::pushingbox}
\begin{figure}
    \centering
    \includegraphics[width=0.485\textwidth]{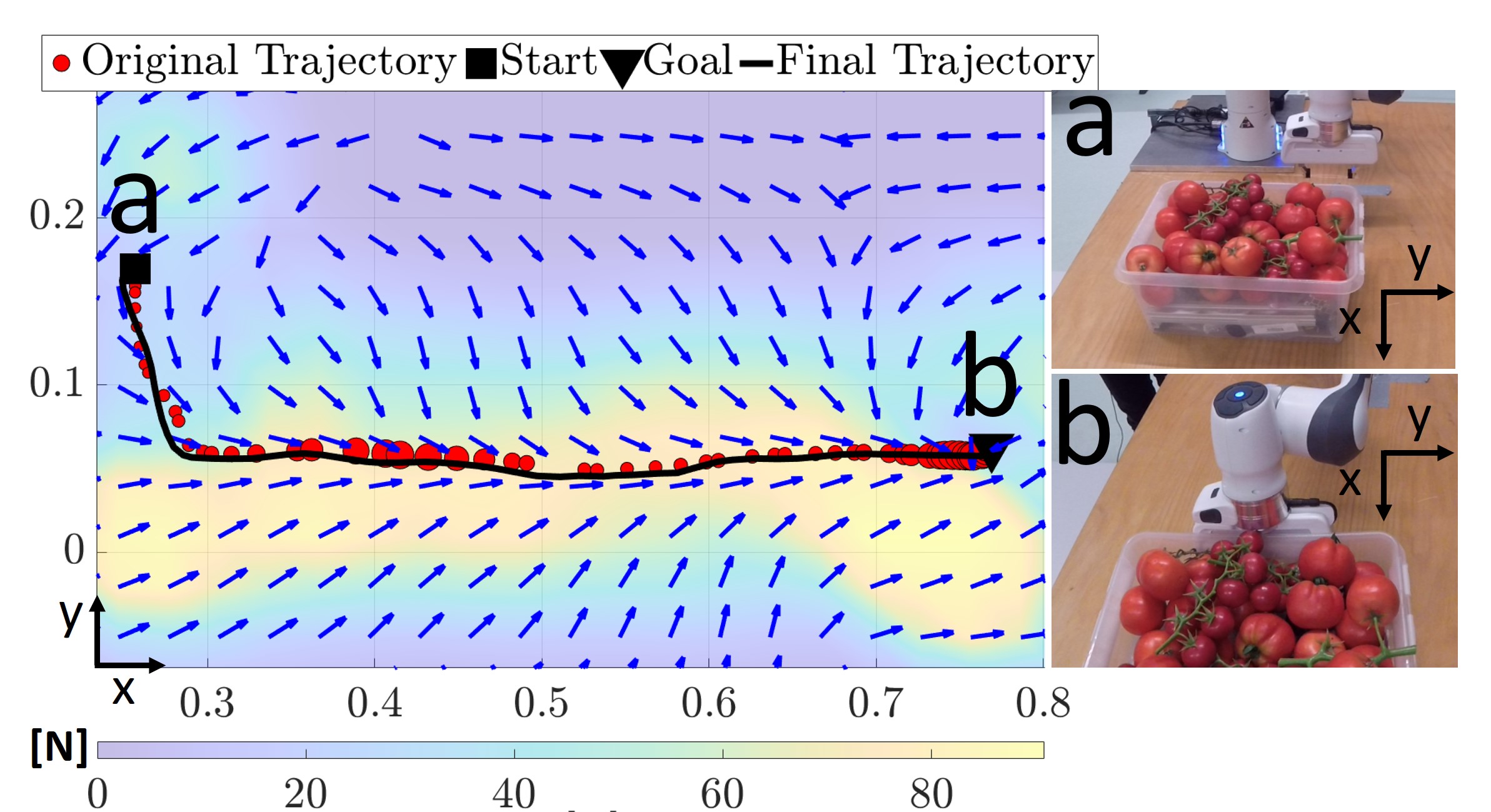}
    \caption{Example of an attractor field for the box pushing task}
    \label{fig:box_pushing}
\end{figure}
In real-world scenarios it can easily happen that objects are unwieldy for the robot's gripper, such that the only remaining option for manipulating said objects is pushing them. In such an interaction, it can happen that the object is removed prematurely, resulting in an unexpected contact loss for the robot. ILoSA enables the limitation of the observed velocities in such a case by limiting the maximum attractor distance. In contrast, if the force were to be achieved by increasing the attractor distance while maintaining a low constant stiffness, these velocities could be noticeably greater.
To verify this, the task was first learned with ILoSA, wherein both stiffness and attractor distance are variable; afterward, it was learned with a variation of the ILoSA algorithm, wherein the force is altered only through a variable attractor distance which is left unbounded. In both the cases, once the interaction was learned, the task was executed with the box being removed while it was being pushed.

\begin{figure}
    \centering
    \includegraphics[width=0.45\textwidth,trim=2.5cm 0cm 3.5cm 0cm,clip]{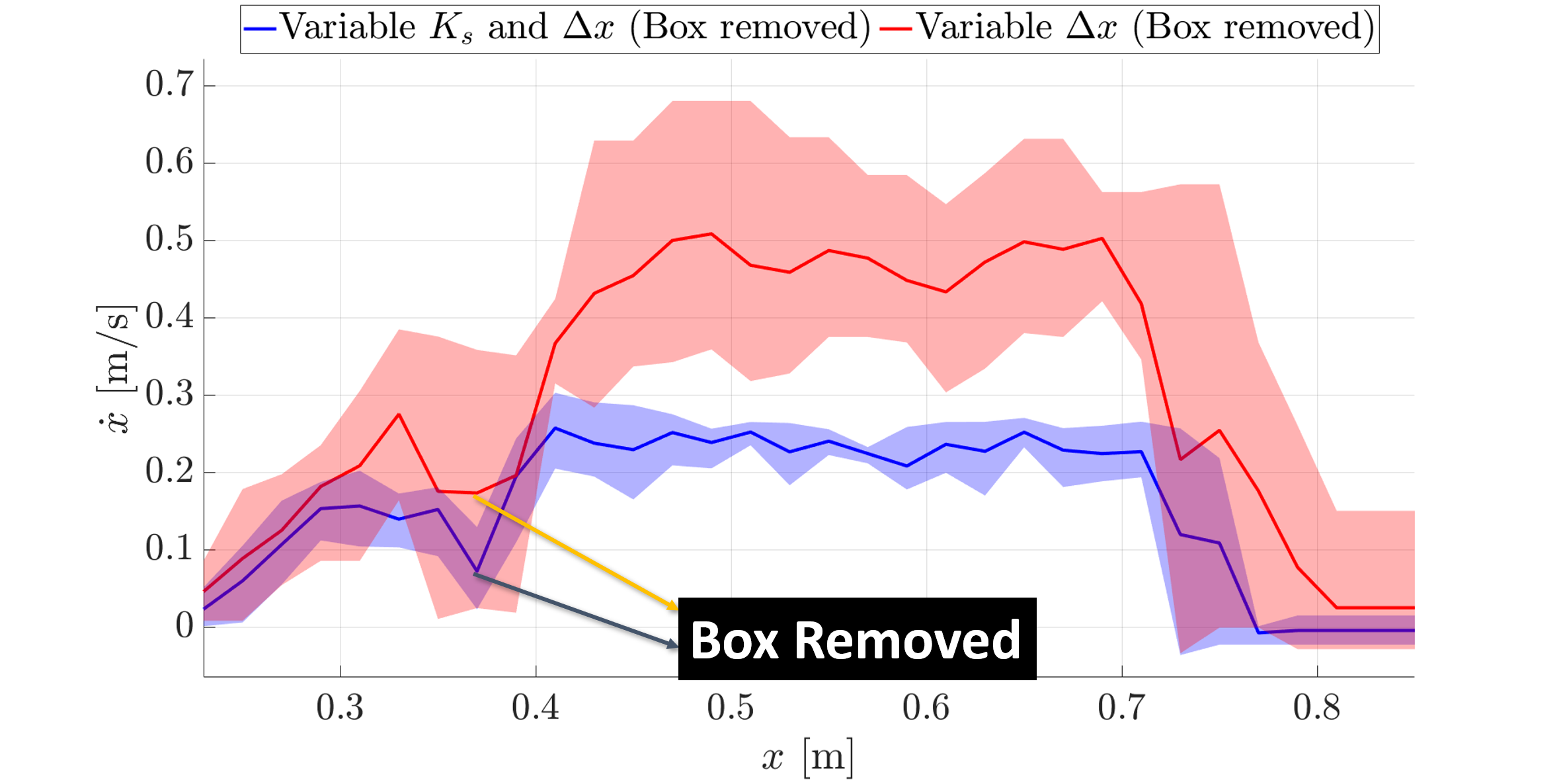}
    \caption{Robot velocity w.r.t. current position along trajectory}
    \label{fig:box_removed}
\end{figure}

Fig.~\ref{fig:box_removed} displays the resulting velocities when varying solely the attractor distance (red), as opposed to concurrently varying the attractor distance and stiffness (blue). As can be seen, the peak velocity for the combined variation of both stiffness $\bm{K_s}$ and attractor distance $\Delta \bm{x}$ is less than half compared to only varying the attractor distance. This in turn allows the limitation of potential impact forces should a person cross the robot's trajectory. 

In terms of performance, ILoSA achieves comparable results to those observed in unplugging, both in terms of error from the goal and data efficiency. The overall correction duration as well as the correction duration relative to the one of the demonstration are larger than those observed when unplugging. This is, however, primarily attributed to the fact that the box pushing scenario has a larger portion of the trajectory in which force has to be applied, in turn requiring more user corrections. An example attractor field can be seen in Fig.~\ref{fig:box_pushing}. On the matter of data efficiency, it should be noted that both in the case of unplugging and in the case of the box, a data efficiency of 100\% was not achieved due to the goal conditioning, which adds the marked goal to the database. For an overview on the task performance, refer to Table~\ref{tab:box}.

\begin{table}[t]
    \caption{Performance in Pushing a Box}
    \centering
    \begin{tabular}{|c||c||c||c||c|}
        \hline
         & \textbf{ Demo \scriptsize[s]} & \textbf{ Fdbk \scriptsize[s]} &  \textbf{ Eff. \scriptsize[\%]} & \textbf{Goal Err. \scriptsize[m]} \\
         \hline
        \textbf{Max} & 6.80 & 4.53 & 98.57 & 0.016 \\
        \hline
        \textbf{Mean} & 5.23 & 4.16 & 95.82 & 0.008 \\
        \hline
        \textbf{Min} & 4.47 & 3.87 & 90.00 & 0.001 \\
        \hline
    \end{tabular}
    \label{tab:box}
\end{table}

\subsection{Cleaning a Whiteboard}
\begin{figure}[t]
    \centering
    \includegraphics[width=\linewidth]{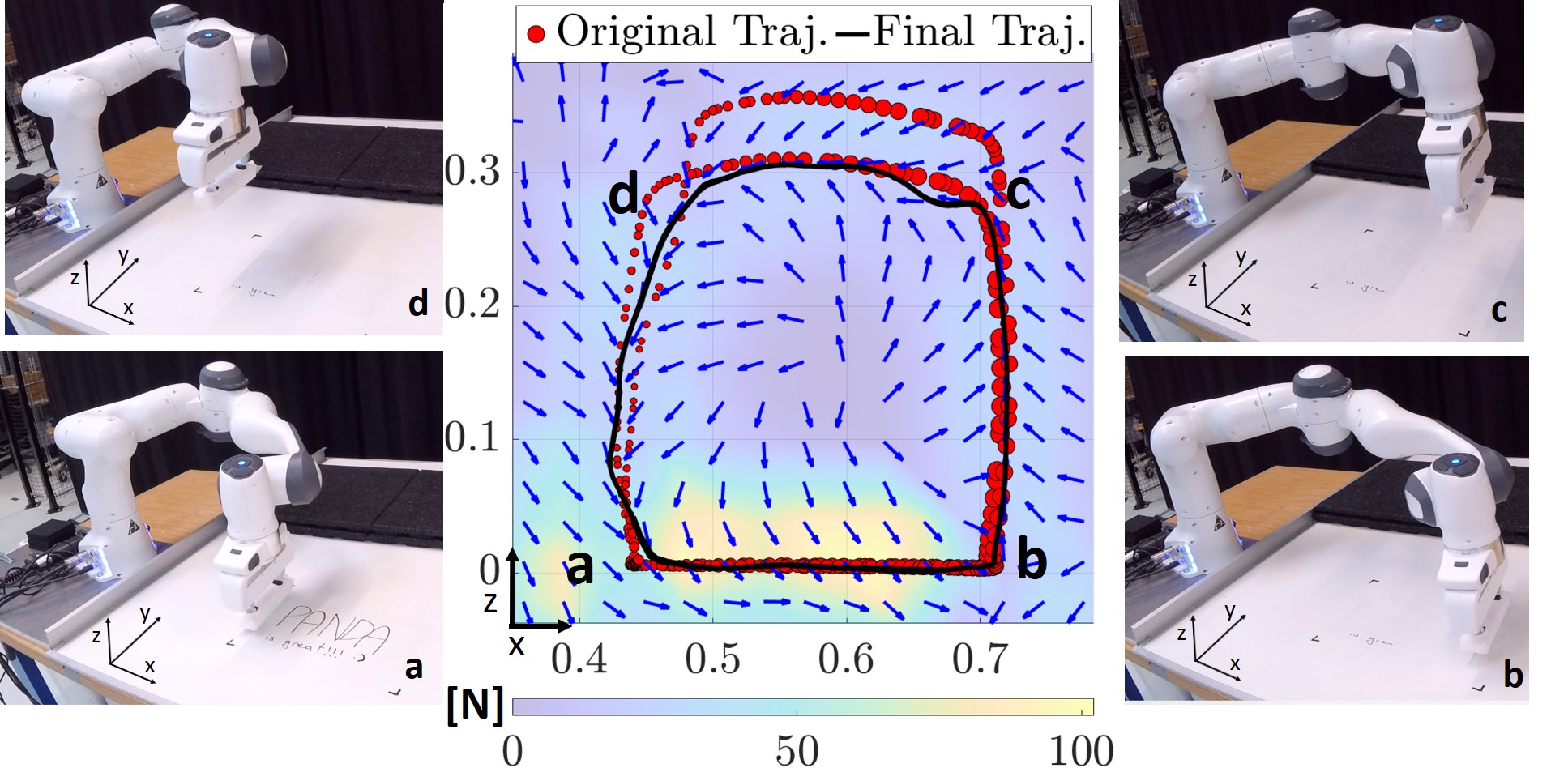}
    \caption{Example of an attractor field for wiping a board}
    \label{fig:board}
\end{figure}
\label{sec::cleanwhiteboard}

In this task, the robot is taught to ensure a whiteboard remains clean and to sustain the movement until the controller is stopped. For this, it is highly desirable that at the end of each operation cycle, the robot returns to the same joint configuration; this property is known as “cyclicity” of motion \cite{chiaverini2016redundant}. For a redundant robot, it is possible for the end-effector to return to the same task-space position and yet the robot to be in a completely different joint configuration. This would result in a feeling of unpredictability for the human watching with consequent frustration. In fact, this is generally the result obtained when methods based on Cartesian impedance control are used to control the robot motion. For solving this problem we also learned a null-space control policy (always from demonstrations) and had it running during the normal execution of ILoSA. During the kinesthetic demonstration it assists the user, allowing them to focus only on the motion of the end effector, and during policy execution it guarantees the cyclicity of the operations.
However, because the main focus was not about the learning of null-space constraints, more details and investigations will be added in future works. 
%However, as this was not the main focus, this addition was not thoroughly tested. 

The execution of this task was deemed successful if the desired area of the board was wiped clean after each loop and the motion continued for at least 5 loops. An example of the resulting attractor field can be seen in Fig.~\ref{fig:board}. 

On the quantitative side, not only were the 5 loops executed successfully, but the robot also remained highly consistent in its motion. 
The consistency was measured as the highest RMSE between each pair of the five loops and it amounted on average to \SI{0.36}{\cm}. Part of the success of a repeatable motion can be credited to the null-space control that ensures a cyclic joint configuration and successively a consistent Cartesian mass matrix and dynamics. In terms of correction time, %once more 
only a short period was spent providing inputs, with an average time of \SI{4.81}{\second}. Out of these inputs, the majority resulted in modifications of the database attaining an average data efficiency of 98.54\%. Additional details are provided in Table~\ref{tab:boardOrigvsAdapt}.

\begin{table}[t]
    \caption{Performance in Cleaning a Whiteboard}
    \centering
    \setlength\tabcolsep{5pt}
    \begin{tabular}{|c||c||c|c||c|c||c|c|}
        \hline
         & \multirow{2}{*}{\makecell{\textbf{ Demo} \\ \textbf{[s]}}} &\multicolumn{2}{c||}{\textbf{Fdbk [s]}} & \multicolumn{2}{c||}{\textbf{Data Eff. [\%]}} & \multicolumn{2}{c|}{\textbf{ Consist. [m]}}\\
         \cline{3-8}
         & & Orig. & Adap. & Orig. & Adap. & Orig. & Adap. \\
         \hline
        \textbf{Max} & 18.27  & 6.4 & 8 & 100 & 91.67 & 0.004 & 0.004 \\
        \hline
        \textbf{Mean} & 16.81 & 4.81 & 5.57 & 98.54 & 83.14 & 0.003 & 0.003 \\
        \hline
        \textbf{Min} & 15.20 & 3.53 & 3.27 & 94.51 & 73.47 & 0.003 & 0.003 \\
        \hline
    \end{tabular}
    \label{tab:boardOrigvsAdapt}
\end{table}
\sidecaptionvpos{figure}{c}

\begin{SCfigure}[][t]
    \centering
    \includegraphics[width=0.56\linewidth, trim = 0 0.15cm 0cm 0.1cm,clip]{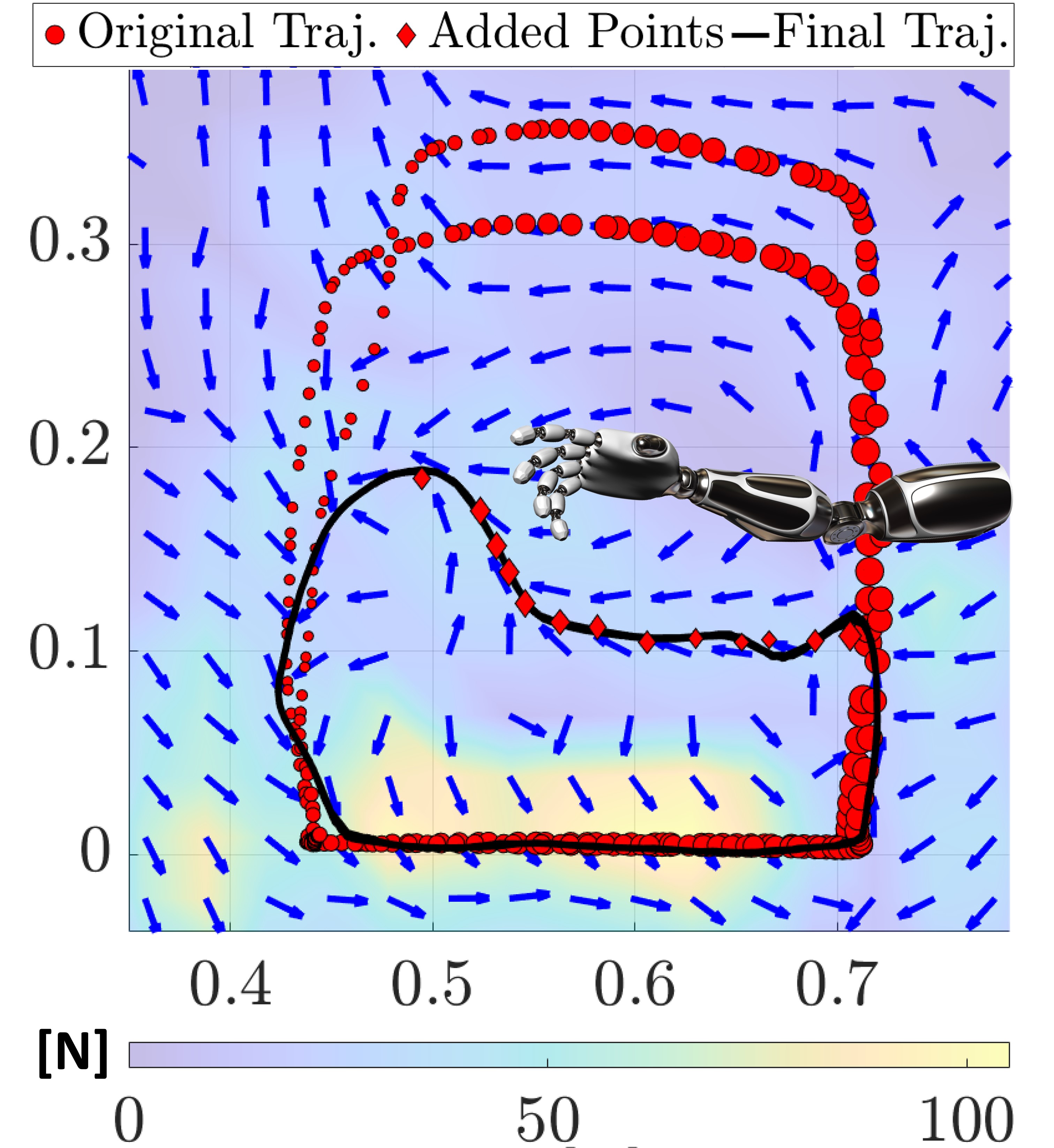}
    \caption{Vector-field and the 5 overlapping final trajectories (black line) after user corrections for learning to avoid the obstacle represented by the arm in the picture.}
    \label{fig::obstacle_avoidance}
\end{SCfigure}

When correcting the original trajectory, adaptations for avoiding the obstacle could be carried out in all five trials. With this we wanted to show how, when close to the original provided samples, the corrections can effectively modify the database to address the desired behaviour and how, when outside the region of certainty, new samples are added, allowing to shape the stabilization field around them. In Fig.~\ref{fig::obstacle_avoidance}, it is possible to see how around the new added points the force field would reject the disturbances and stay on the new desired trajectory. Through an additional \SI{5.57}{\second} of corrections on average, and with an average data efficiency of 83.14\%, the motion was successfully adapted, resulting in attractor fields similar to the one seen in the figure.

\subsection{Evaluation with Non-Expert Users: Plugging}
\label{sec::plugging}
% \todo[inline]{We need to mention the questionaire LTX and VAN DER LAAN}
After observing that the framework was capable of remaining in close vicinity of the demonstrated trajectory, the task of plugging was taken for its clear constraint on the goal position and the requirement to apply force in order to successfully accomplish the task.

In order to verify that the successful completion of the task was not dependent on executing the training in a specific manner, non-expert participants were asked to carry out the experiment. A total of fifteen participants aged 23~-~32 took part in an additional engineering validation, rather than a dedicated human factors study. The participants were allowed to perform the task twice. The first time was to give participants the opportunity to get familiar with the teleoperation device for a couple of minutes. The second time was then treated as the official trial, wherein the participants provided a new demonstration and were allowed to provide as many rounds of correction as needed until the robot carried the task out successfully.

\begin{table}[t]
    \caption{Performance in Inserting a Plug}
    \centering
    \begin{tabular}{|c||c||c||c|}
        \hline
         & \textbf{ Fdbk \scriptsize[s]} & \textbf{ Total Time \scriptsize[s]} &  \textbf{ Rounds of Correction} \\
         \hline
        \textbf{Max} & 17.49 & 193.85 & 6 \\
        \hline
        \textbf{Mean} & 4.72 & 95.84 & 2.07 \\
        \hline
        \textbf{Min} & 1.22 & 54.46 & 1 \\
        \hline
    \end{tabular}
    \label{tab:participants}
\end{table}

All participants were able to complete the task in a fairly short period of time. On average, the total time needed to train the task was about one and a half minutes, which includes both the time needed for providing the kinesthetic demonstration as well as any amount of correction rounds needed for the robot to successfully insert the plug. A summary of the performance metrics can be viewed in Table \ref{tab:participants}. In terms of data efficiency, all of the participants completed the task with 100\% data efficiency, indicating that the robot never left the region of the demonstration.

Additionally, participants were asked to fill out the NASA TLX and Van der Laan questionnaires after completion of the task. A majority of the participants reported low mental demand and found the teaching method overall helpful and easy to use.

\section{Conclusions and Future Work}
\label{conclusion}

We have introduced a non-parametric and interactive approach to learning different types of force interaction tasks from humans, while exploiting impedance control to ensure safe interaction. All aspects of the interactions, from the attractor distance and stiffness at the end-effector to the null-space control were successfully modelled with the help of Gaussian Processes. This enables non-experts to create complex robotic interaction skill that would alternatively require a set of expert skills through manual programming.

Making use of the learned model parameters, it was possible to establish two additional safety features. The first is the reduction of the stiffness should the robot be too far from the demonstrated region, eventually bringing it to a halt. The second is a stabilisation prior, which helps to steer the robot back to the closest area with low variance, consequently returning it back to the demonstrated region. As a result, this enables the rejection of disturbances. 

Moreover, the stabilisation prior was able to infer that its effect should be reduced in the areas between demonstrations, allowing the robot more freedom of movement in those areas. However, in the case of desired multi-modal behaviours, e.g., for obstacle avoidance, this could be obtained with a constraint on the maximum lengthscale of the used kernel. This is equivalent to the generation of multiple separate variance furrows rather than a single wider one.

Our investigations further showed that the ILoSA framework can be implemented  for carrying out both goal-oriented and periodic movements. When used in combination with a null-space control, which enabled cyclicity, a high consistency of the motion was attained. Overall, ILoSA exhibited good reliability in the execution of the examined tasks, while learning in a user-friendly and data efficient manner.

Due to the successful applications in the force tasks in which it was tested, ILoSA will be extended to further challenges in the field of robot manipulation. The learning will not only focus on a single trajectory but on the assembly of movement sequences for more complex tasks, always learning from demonstration. Adaptation of the motion with respect to a particular reference frame in each segment will be investigated using human feedback and the information on the model confidence for solving possible ambiguity as in \cite{Franzese2020CoRL}.
Modulation of the velocity from corrections will be further investigated for learning tasks such as fast grasping. 
In the future, we will also study different modalities of feedback, such as force feedback from haptic devices or kinesthetic perturbations, along the lines of \cite{kronander2013learning}. The goal is to prove the possible leap from academic to industrial/household applications. 

\bibliographystyle{IEEEtran}
\bibliography{biblio}

\end{document}